\documentclass{ieeeaccess}
\usepackage{cite}
\usepackage{amsmath,amssymb,amsfonts}
\usepackage{algorithmic}
\usepackage{graphicx}
\usepackage{textcomp}
\def\BibTeX{{\rm B\kern-.05em{\sc i\kern-.025em b}\kern-.08em
    T\kern-.1667em\lower.7ex\hbox{E}\kern-.125emX}}
    
\usepackage{booktabs}
\usepackage{tabularx}
\usepackage{hyperref}
\usepackage{eucal}
\usepackage{multirow}
\usepackage[bottom]{footmisc}
\usepackage{todonotes}

\usepackage{tikz}
\NewSpotColorSpace{PANTONE}
  \AddSpotColor{PANTONE} {PANTONE3015C} {PANTONE\SpotSpace 3015\SpotSpace C} {1 0.3 0 0.2}
  \SetPageColorSpace{PANTONE}%
  
\usepackage{orcidlink}

\usepackage{subfigure}

\def\BibTeX{{\rm B\kern-.05em{\sc i\kern-.025em b}\kern-.08em
    T\kern-.1667em\lower.7ex\hbox{E}\kern-.125emX}}
\begin{document}
\history{Date of publication xxxx 00, 0000, date of current version xxxx 00, 0000.}
\doi{10.1109/ACCESS.2017.DOI}

\title{I-AID: Identifying Actionable Information from Disaster-related Tweets}
\author{\uppercase{Hamada M. Zahera}\authorrefmark{1,2}\authorrefmark{\orcidlink{0000-0003-0215-1278}}, 
\uppercase{Rricha Jalota}\authorrefmark{1}\authorrefmark{\orcidlink{0000-0003-1517-6394}
}, 
\uppercase{Mohamed Ahmed Sherif}\authorrefmark{1}\authorrefmark{\orcidlink{0000-0002-9927-2203
}}
and \uppercase{Axel-Cyrille Ngonga Ngomo}\authorrefmark{1}\authorrefmark{\orcidlink{0000-0001-7112-3516
}}}
\address[1]{DICE group, Department of Computer Science, Paderborn University, Germany}
\address[2]{Faculty of Computers and Information, Menoufia University, Egypt}

\tfootnote{This work has been submitted to the IEEE for possible publication. Copyright may be transferred without notice, after which this version may no longer be accessible.}

\markboth
{Zahera \headeretal: Preparation of Papers for IEEE TRANSACTIONS and JOURNALS}
{Author \headeretal: Preparation of Papers for IEEE TRANSACTIONS and JOURNALS}

\corresp{Corresponding author: Hamada M. Zahera (hamada.zahera@uni-paderborn.de)}

\begin{abstract}
Social media plays a significant role in disaster management by providing valuable data about affected people, donations and help requests.
Recent studies highlight the need to filter information on social media into fine-grained content labels. 
However, identifying useful information from massive amounts of social media posts during a crisis is a challenging task.
In this paper, we propose I-AID, a multimodel approach to automatically categorize tweets into multi-label information types and filter critical information from the enormous volume of social media data. 
I-AID incorporates three main components: 
i) a BERT-based encoder to capture the semantics of a tweet and represent as a low-dimensional vector,
ii) a graph attention network (GAT) to apprehend correlations between tweets' words/entities and the corresponding information types, and 
iii) a \textit{Relation Network} as a learnable distance metric to compute the similarity between tweets and their corresponding information types in a supervised way.
We conducted several experiments on two real publicly-available datasets. 
Our results indicate that I-AID outperforms 
state-of-the-art approaches in terms of weighted average F1 score by
$+6\%$ and $+4\%$ on the TREC-IS dataset and COVID-19 Tweets, respectively. 
\end{abstract}

\begin{keywords}
Crisis Information, Contextualized Text Embedding,  Social Media Analysis, Graph Attention Network, Meta Learning.
\end{keywords}

\titlepgskip=-15pt

\maketitle

\section{Introduction}
\label{sec:introduction}
Social media has become a key medium for sharing information during emergencies~\cite{DBLP:journals/pacmhci/ZadeSRK0S18}. 
The major difference between social media and traditional news sources is the possibility of receiving feedback from affected people in real time. 
Relief organizations can benefit from this two-way communication channel to inform people and gain insights from situational updates received from affected people.
Hence, extracting crisis information from posts on social media (e.g., tweets) can substantially leverage situational awareness and result in faster responses. 

Most previous works~\cite{to2017identifying,stowe2018improving} addressed information extraction from social media as a binary text classification problem (e.g., with the labels \texttt{Relevant} and \texttt{Irrelevant}). 
However, there is a lack of efficient systems that can map relevant posts to more fine-grained labels as, for example, defined in~\cite{DBLP:conf/iscram/McCreadieBS19} (see Figure~\ref{fig:tweetexample}). 
Such fine-grained labels are particularly valuable for crisis responders as they filter critical information to deliver disaster responses quickly. 
In particular, labeling disaster-related tweets using multiple labels allows the rapid detection of tweets with \textit{actionable} information. 
Table~\ref{tab:informationTypes} shows the list of information types (which we use as labels) defined by~\cite{DBLP:conf/iscram/McCreadieBS19}. 
We adopt the definition of \textit{actionable} tweets as formalized in~\cite{DBLP:journals/pacmhci/ZadeSRK0S18}.
\textit{Actionable} tweets are defined as the ones that would generate an immediate alert for individuals (i.e., stakeholders) responsible for the information type with which they are labeled (e.g., \texttt{SearchAndRescue}, \texttt{MovePeople}). 
This stands in contrast to  \textit{non-actionable} tweets that are labeled with labels such as \texttt{Hashtags} or \texttt{FirstPartyObservation} (see Table~\ref{tab:informationTypes}).

\Figure[t!] (topskip=0pt, botskip=0pt, midskip=0pt)[width=0.95\columnwidth]{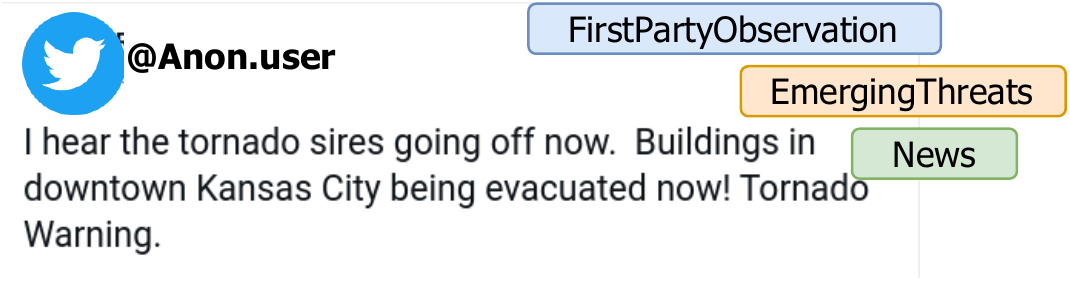}{Example of multi-label tweet classification with assigned labels: FirstPartyObservation, EmergingThreats and News.     \label{fig:tweetexample}}

On the other hand, categorizing tweets is known to be a challenging short text Natural Language Processing (NLP) task~\cite{song2014short}.
This is because tweets 
i) does not possess sufficient contextual information, and 
ii) is inherently noisy (e.g., contains misspellings, acronyms, emojis, etc.). 
Moreover, in the multi-label case, the classification task becomes even more challenging because a tweet can belong to one or more labels simultaneously. 
	
In this paper, we aim to 
i) label disaster-related tweets with fine-grained information types so as to 
ii)  identify \textit{actionable} or \textit{critical} tweets that might be relevant for disaster relief and support disaster mitigation. 
Our approach contains three components: 
First, we use BERT as a sentence encoder to capture the semantics of tweets and to represent them as low-dimensional vectors. 
Second, we employ a graph attention network (GAT) to capture correlations between the words and entities in tweets and the labels of said tweets. 
Finally, we use a \textit{Relation Network}~\cite{sung2018learning} as a learnable distance metric to compute the similarity between the vector representation of tweets (obtained from the BERT encoder) and the vector representation of labels (obtained from the GAT) in a supervised way. 
By these means, our system integrates a contextualized representation of tweets with correlations between tweets and their labels. 
The main contributions of this paper can be summarized as follows:
	\begin{itemize}
		\item We propose a multimodel 
		approach (dubbed I-AID) to categorize disaster-related tweets into multiple information types. 
		\item Our approach leverages a contextualized representation from a pretrained language model (BERT) to capture the semantics of tweets.  
		In addition, our approach employs a GAT component to capture the structural information between the words and entities in tweets and their labels. 
		\item We employ a \textit{learnable} distance metric, in a supervised way, to learn the similarity between a tweet's vector and the labels' vectors.  
		\item We conduct several experiments to evaluate the performance of our approach and state-of-the-art baselines in multi-label text classification. 
	\end{itemize}
	
The rest of this paper is structured as follows:
In Section~\ref{sec:relatedworks}, we discuss previous work on the classification of crisis information on social media. 
In Section~\ref{sec:approach}, we describe the preliminaries and architecture of our proposed approach. 
Finally, we discuss the experimental results in Section~\ref{sec:experiment} and conclude the paper in Section~\ref{sec:conclusion}. 

	
\begin{table*}[h!]
	\centering 
	\small
	\caption{Crisis Information Types (i.e., Labels or Classes)\label{tab:informationTypes}}
	\begin{tabularx}{0.85\textwidth}{@{}p{2.1cm}p{3cm}p{7cm}@{}}
		\toprule
		\textbf{Intent Type }& \textbf{Information Type} & \multicolumn{1}{c}{\textbf{Description}} \\ \midrule
		
		\multirow{3}{*} {REQUEST}& GoodServices & Request for a particular service or physical good \\
		& SearchAndRescue & The user is requesting a rescue for themselves or others\\
		& InformationWanted & The user is requesting information \\ \midrule
		\multirow{11}{*}{REPORT} & Weather & Weather report \\
		& FirstPartyObservation & The user is giving an eyewitness account \\
		& ThirdPartyObservation & The user is reporting information from someone else\\
		& EmergingThreats & Problems that cause loss or damage \\
		& ServiceAvailable & Someone is providing a service \\
		& SignificantEventChange & New occurrence to which officers need to respond\\
		& MultimediaShare & Shared images or video \\
		& Factoid & The user is reporting some facts, typically numerical \\
		& Official & Report by a government or public representative \\
		& CleanUp & Report of the cleanup after an event \\
		& Hashtags & Report with hashtags correspond to each event \\ \midrule
		\multirow{3}{*}{CALLTOACTION} & Volunteer & Call for volunteers to help in response efforts \\
		& Donations & Call for donations of goods or money \\
		& MovePeople & Call to leave an area or go to another area \\ \midrule
		\multirow{6}{*}{OTHER} & PastNews & The post is reporting an event that has occurred \\
		& ContinuingNews & The user is providing/linking to a continuous event\\
		& Advice & Provide some advice to the public\\
		& Sentiment & The post is expressing some sentiments about an event\\
		& Discussion & Users are discussing an event\\
		& Irrelevant & The post is irrelevant\\
		\midrule
	\end{tabularx}
\end{table*}

\section{Related Work}\label{sec:relatedworks}
The objective of this work is to categorize disaster-related tweets into multiple information types.
Therefore, we relate our work to extract \textit{disaster-related information on social media}, \textit{multi-label text classification} and \textit{meta learning}. In the following, we briefly discuss the state of the art in each of these areas. 
	
	\subsection{Extracting Disaster-related Information from Social Media}
	Several studies demonstrate the role of social media as a primary source of information during disasters~\cite{cresci2015crisis}. 
	While some works~\cite{PalshikarAP18} focused on filtering relevant information from tweets,  
	others (e.g., \cite{olteanu2015expect,mccreadie2019trec}) proposed annotation schemes to classify tweets into fine-grained labels that consider the attitude, information source and decision-making behavior of people tweeting before, during and after disasters. 
	To advance the state of social media crisis monitoring solutions, initiatives like~\cite{imran2016lrec}  
	have been rolled out in recent years.
	One of them is the Incident Streams (TREC-IS) track~\cite{mccreadie2019trec} of the Text REtrieval Conference, which commenced in 2018. 
	The track aims to categorize disaster-related tweets into multiple information types. 
	In this work, we study the TREC-IS dataset and adopt the definition of \textit{actionable} information from the authors of the TREC-IS challenge. In addition, we employ their performance metric (called Accumulated Alert Worth~\cite{TRECIS-AAW}) to evaluate our system in identifying \textit{actionable} information in tweets.
	
	\subsection{multi-label Text Classification}
	Earlier works in text classification~\cite{sriram2010short} consider feature engineering and model training as different subtasks. 
 With the advent of end-to-end deep learning approaches~\cite{miyazaki2019label} and the attention mechanism~\cite{vaswani2017attention}, there has been a significant advancement in the field of multi-label text classification. 
Pretrained language models (e.g, BERT~\cite{devlin2018bert}) are becoming increasingly popular for text classification~\cite{miyazaki2019label}.
However, since BERT only captures the local contextual information, the BERT embeddings do not sufficiently capture the global information about the lexicon of a language.~\cite{DBLP:conf/ecir/LuDN20} 
To circumvent this and comprehend the global relations among words in a vocabulary, graph-based approaches such as graph convolution network (GCN)~\cite{kipf2016semi} and graph attention network (GAT)~\cite{yao2019graph} have been promising.
	
Recent studies~\cite{DBLP:conf/ecir/LuDN20,DBLP:conf/icaart/PalSS20} have exploited the advantages of combining BERT and graph networks. 
In VGCN-BERT~\cite{DBLP:conf/ecir/LuDN20}, a GCN is used to capture the correlation between words at the vocabulary level (i.e., global information). 
For instance, given a vocabulary, the GCN would relate the meaning of \textit{"new"} to \textit{"innovation"} and \textit{"exciting"}, similar to context-independent word embeddings like \textit{word2vec}~\cite{DBLP:journals/corr/abs-1301-3781}. For an input sentence, the \textit{local} contextual information is captured using BERT embeddings, while the global information pertaining to words in a sentence is extracted using graph embeddings and subsequently concatenated with BERT. The two representations of BERT and GCN then interact via the self-attention mechanism to perform the classification task.

In a similar work, Ankit Pal~\textit{et al.}~\cite{DBLP:conf/icaart/PalSS20} leverage the combination between BERT embeddings and GAT to learn feature representation for text in a multi-label classification task. 
Their proposed approach (dubbed MAGNET) employs two components: First, a BiLSTM network with BERT embedding is used to capture text representation into an embedding vector. In the second component, the authors use GAT to learn a feature vector for labels. In particular, their GAT models the correlation between words and the corresponding labels, then averages the labels' vectors into a single output vector. Finally, the authors use a dot-product function to compute the similarity between the input's vector from BiLSTM and the label's vector. 
In contrast, our approach differs from both MAGNET and VGCN-BERT in computing the similarity between a tweet's representation and its labels' vectors. 
We employ a GAT model to explicitly infuse the correlation information of the entities and labels of a tweet with the tweet's contextualized BERT representation. 
While MAGNET and VGCN-BERT use either a fixed and linear distance metric (dot-product function) or self-attention to measure the similarities, our approach benefits from a deeper end-to-end neural architecture to learn this distance function. In particular, we employ meta-learning to learn 
the mapping between the input features and multi-label output in a supervised way. 
	
	
	\subsection{Meta Learning}
	Meta learning (also called \textit{learning-to-learn} paradigm) refers to the process of improving a learning algorithm over multiple learning episodes.
	In contrast to conventional machine learning approaches, which improve model prediction over multiple data instances, the meta-learning framework treats tasks as training examples to solve a new task~\cite{DBLP:journals/corr/abs-2004-05439}. 
	In our study, we employ a specific branch of meta learning called metric learning. 
	Metric learning learns a distance function between data samples so that the test instances get classified by comparing them to the labeled examples. 
	The distance function consists of 
	i) an embedding function, which encodes all instances into a vector space, and 
	ii) a similarity metric, such as cosine similarity or Euclidean distance, to calculate how close two instances are in the space~\cite{DBLP:journals/corr/abs-2007-09604}. 
	Recently, many approaches have been developed to perform this task, such as \textit{Siamese}~\cite{koch2015siamese}, \textit{Matching}~\cite{vinyals2016matching}, \textit{Prototypical}~\cite{snell2017prototypical}, and \textit{Relation Network}~\cite{sung2018learning}. 
	While the embedding function in all of these approaches is a deep neural network, they differ in terms of the similarity function. 
	Unlike its predecessors, which rely on a fixed similarity metric (such as cosine, Euclidean, etc.),  Relation Network employs a flexible function approximator to learn similarity and focuses on learning a good similarity metric in a supervised way. 
	The use of function approximators eliminates the need to manually choose the right metric (e.g., Euclidean, cosine, Manhattan). 
	By jointly learning the embedding and a nonlinear similarity metric, Relation Network can better identify matching/mismatching pairs~\cite{garcia2018fewshot}. 
	For this purpose, we use the Relation Network in our work for learning the similarity metric.

\section{Our Approach}\label{sec:approach}

\begin{table}[tb]
\small
\centering
\caption{A List of Symbols Used in This paper.}
\begin{tabularx}{\columnwidth}{@{}p{1.2cm}p{6.5cm}XX@{}}
\toprule
\textbf{Symbol} & \textbf{Description} \\ \hline
 $S$ & Number of tweets in the dataset.\\
 $w$ & Tweet tokens (e.g., word or entity). \\
 $y^{(i)}$ & Ground-truth multi-label assigned to a tweet $i$.\\
 $\hat{y}^{(i)}$ & Predicted multi-label assigned to a tweet $i$.\\ 
 $\lambda_i$ & A single label/information type for a tweet.\\ 
 \midrule
 $N$ & Number of nodes in a graph \\
 $\mathcal{V}$& Nodes of a graph \\
 $\mathcal{E}$ & Edges between nodes in a graph\\
 $A$ & Adjacency matrix of a graph \\
 $\tau^{(i)}$ & Embedding vector of tweet $i$ learned by BERT\\
 $h^{(i)}$ & Embedding vector of node $v^{(i)}$\\
 \midrule
 $F$ & Dimension of node vector \\
 $\iota^{(i)}$ & Embedding vector for label $\lambda^{(i)}$\\
 $Z$ & The concatenated vector of $\tau^{(i)}$ and  $\iota$ \\
 $\mathcal{L}$& Binary cross-entropy loss function\\
 $\alpha_{ij}$ & Attention score between nodes $v^{(i)}$ and $v^{(j)}$\\
 \midrule
 $hPW(t)$ & Scoring function for high priority tweets.\\
 $hPW(l)$ & Scoring function for low priority tweets.\\
 \bottomrule
\end{tabularx}
\end{table}

\Figure[t!] (topskip=0pt, botskip=0pt, midskip=0pt)[width=0.95\textwidth]{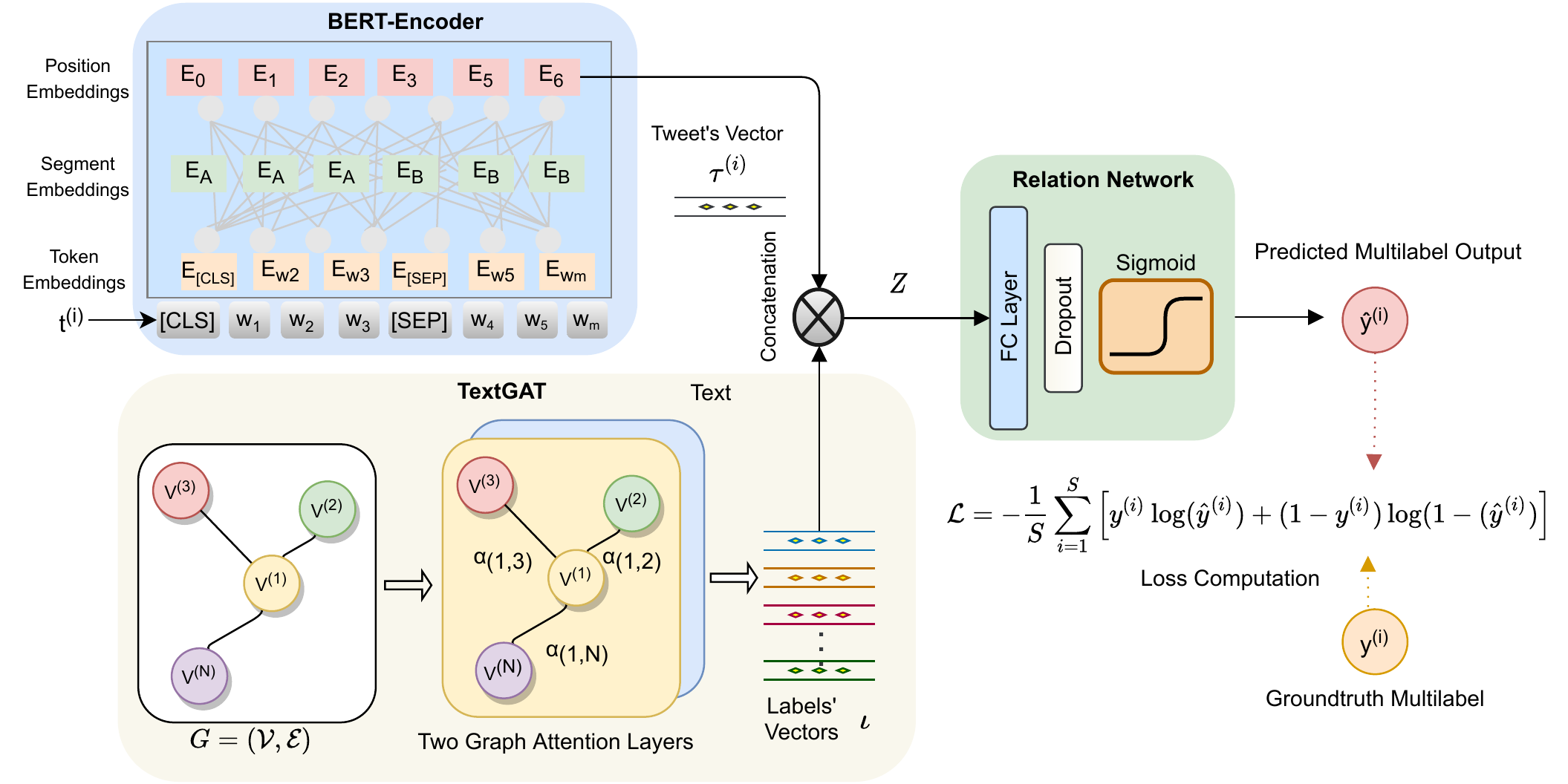}
		{The I-AID architecture: BERT-Encoder embeds tweet $t^{(i)}$ into a feature vector $\tau^{(i)}$. TextGAT builds a graph $G$ from our dataset, employs graph  attention layers and output labels vectors $\iota$. Relation Network learns a distance metric between $\tau^{(i)}$ and $\iota$, then outputs predicted labels $\hat{y}^{(i)}$ for $t^{(i)}$.
		\label{fig:I-AID}
	}
	
We begin this section by giving a formal specification of the multi-label tweet classification problem. 
Afterward, we discuss the details of each component of our approach in Section~\ref{sec:I-AID Architecture}.
Figure~\ref{fig:I-AID} gives an overview of our approach and how its components work together. 
	
	\subsection{Problem Formulation}\label{sec:problem_formulation}
	Let $T$ be a set of tweets and $\Lambda=\{\lambda_1, \lambda_2,\cdots ,\lambda_k\}$ be a set of $k$ predefined labels (also called information types, see Table~\ref{tab:informationTypes}). 
	We formulate the problem of \textit{identifying crisis information from tweets} as a multi-label classification task, where a tweet $t$ can be assigned one or more labels from $\Lambda$ simultaneously. 
	Our task is to learn a multi-label classifier $\mathbf{M}: T \to \{0,1\}^k$ that maps tweets $T$ to relevant labels from $\Lambda$.
	We assume a supervised learning setting where  
	a training data $\mathcal{D}=\big\{ (t^{(i)},y^{(i)})\times \{0,1\}_{j}^k \big\}_{i=1}^S$ consists of $S$ tweets. 
	Hence, each tweet $t^{(i)}$ is labelled with a set of corresponding labels $y^{(i)}$, where $y^{(i)}_j = 1$ means that $t^{(i)}$ belongs to the class $\lambda_j$. 
	Conversely, $y^{(i)}_j = 0$ means that $t^{(i)}$ does not belong to the class $\lambda_j$. 
	The goal of our approach is to learn the function $\mathbf{M}$ by using three neural networks.
	First, we transform tweet $t^{(i)}$ into an embedding vector $\tau^{(i)}$ using a pretrained BERT model. 
	In parallel, our approach learns labels' embeddings $\iota$ using a graph attention network (GAT). 
	These are then concatenated with the tweet embedding $\tau^{(i)}$. 
	Finally, these vectors are fed to our last component (Relation Network) to identify relevant labels for $t^{(i)}$. 
	
	\subsection{The I-AID Architecture}\label{sec:I-AID Architecture}
	
	\subsubsection{BERT-Encoder}
	This is the first component in our system that transforms an input tweet into a vector representation $\tau$ of its contextual meaning. 
	As shown in Figure~\ref{fig:I-AID}, the BERT-Encoder takes tweet $t^{(i)}$ with $m$ tokens $[w^{(i)}_1,w^{(i)}_2,\dots ,w^{(i)}_m]$ and outputs the embedding vector $\tau^{(i)}$. We employ a BERT-base architecture with $12$ encoder blocks, $768$ hidden dimensions, and $12$ attention heads. We refer readers to the original BERT paper~\cite{devlin2018bert} for a detailed description of its architecture and input representation. 
	Furthermore, a special preprocessing is performed for BERT input. A \texttt{[CLS]} token is appended to the beginning of the tweet and another token \texttt{[SEP]} is inserted after each sentence as an indicator of sentence boundary. Each token $w^{(i)}$ is assigned three kinds of embeddings (token, segmentation, and position). These three embeddings are summed to a single output vector $\tau^{(i)}$ that captures the meaning of an input tweet.
	
\subsubsection{Text-Graph Neural Network (TextGAT)}
Traditional methods (e.g., word2vec~\cite{DBLP:journals/corr/abs-1301-3781}) can properly capture features from a text. 
However, these methods ignore the structural information and relationship between words in a text corpus~\cite{peng2018large}. 
The recently proposed graph networks~\cite{yao2019graph} aim to tackle this challenge by modeling text as a graph where words are nodes and relations between them are edges. 
In our work, we build a graph $G=(\mathcal{V}, \mathcal{E})$ from the dataset $\mathcal{D}$, where $\mathcal{V}$ and $\mathcal{E}$ represent nodes set and their edges, respectively. Each node $v^{(i)} \in \mathcal{V}$ can be a \texttt{word, named-entity\footnote{We spot named-entities in tweets using spaCy entity recongnizer \url{https://spacy.io/api/entityrecognizer}}} or \texttt{label} (tweet's class or information type). 
We represent nodes using a feature matrix $\mathbf{H}=\{h^{(1)}, h^{(2)}, \cdots ,h^{(N)}\}$ where $h^{(i)} \in \mathbb{R}^F$ is the feature vector of node $v^{(i)}$ with $F$ dimension and $N$ is the number of nodes. 
First, we initialize the nodes' representation $\mathbf{H}$ with pretrained embeddings from Glove embedding~\cite{pennington2014glove}.
Further, relations between nodes are modeled using an adjacency matrix $\mathbf{A} \in \mathbb{R}^{N\times N}$.

As shown in Figure~\ref{fig:I-AID}, TextGAT component has two graph attention layers.
Each layer takes nodes' features $\mathbf{H}$ as input and performs an \textit{attention} operation~\cite{velivckovic2017graph} to learn a new feature $\mathbf{\hat{H}}=\{\hat{h}^{(1)}, \hat{h}^{(2)}, \cdots ,\hat{h}^{(N)}\}$ for each node based on its neighbours' importance (i.e., \textit{attention from its neighbours}). 
Hence, we employ the shared attention mechanism $att: \mathbb{R}^{\hat{F}} \times \mathbb{R}^{\hat{F}} \longrightarrow \mathbb{R}$ over all nodes. 
The graph attention operated on the node representation can be written as: 
\begin{equation}\label{eq:att}
	    \alpha_{ij}=att~(\mathbf{W}v^{(i)},\mathbf{W}v^{(j)})
\end{equation}

where $att$ is a single-layer feedforward network, parametrized by a weight matrix $\mathbf{W}\in \mathbb{R}^{\hat{F}\times F}$ which is applied to every node. 
Finally, we use a softmax function to normalize the attention scores as shown in Eq.~\ref{eq:softmax}.
	\begin{equation}\label{eq:softmax}
	\alpha_{ij}=\frac{exp(\alpha_{ij})}{\sum_{k\in N_i}exp(\alpha_{ik})}
	\end{equation}

To this end, TextGAT learns the structural information between nodes based on the relative importance of neighbours. 
The learned representations of labels are then extracted and concatenated with the tweet's vector as input for the last component, as shown in Figure~\ref{fig:I-AID}. 
	
\subsubsection{Relation Network.}
\label{relation-network} 
In this component, we aim to learn a similarity metric in a supervised way (also called learning-to-learn or meta learning) between the tweet's vector $\tau^{(i)}$ and labels vectors $\iota$. 
Furthermore, we employ a neural network as a learnable, nonlinear distance function that \textit{learns} how to match similarity (i.e., relation) between the tweet's vector and each label. Relation Network takes as input the concatenated matrix $Z= \tau^{(i)} \otimes \iota$ of BERT-Encoder output with the labels' vectors. Since our task is multi-label classification, we use the binary cross-entropy as a loss function in Eq.~\ref{eq:binary_cross_entropy}. Then we use a sigmoid function in the output layer to compute the probability of each label independently over all possible labels ($\Lambda$), in contrast to a softmax function which only considers the label with highest probability. Finally, a set of relevant labels is returned as a final output of our approach.
\begin{equation}
	\label{eq:binary_cross_entropy}
	\mathcal{L}=-\frac{1}{S}\sum_{i=1}^{S}\left[y^{(i)}\log(\hat{y}^{(i)})+(1-y^{(i)})\log(1-(\hat{y}^{(i)})\right]
	\end{equation}
	
	where $y^{(i)}$ and $\hat{y}^{(i)}$ are the predicted and ground-truth labels of tweet $i$ respectively. $S$ is size of tweets in the training dataset. 

\section{Experiments}\label{sec:experiment}
In this section, we report the evaluation results of our approach and baseline methods. 
We aim to answer the following research questions:
	\begin{itemize}
	    \item[$Q_1$:]
	    How does our approach perform compared with  state-of-the-art multi-label models in short text (e.g., tweets) classification?
	    \item[$Q_2$:] How effective is our approach in identifying tweets with \textit{actionable} information?
	    \item[$Q_3$:] How does each component in our approach affect the overall performance (i.e., Ablation Study)?
	\end{itemize}
		
	
\begin{table}[ht]
\centering
\caption{Overview of the Datasets.}
\begin{tabular}[t]{lcccc}
\toprule
\textbf{Datasets} & \textbf{\# Train} & \textbf{\# Valid} & \textbf{\# Test} & \textbf{\# Classes}\\
\midrule
TREC-IS &  27,467 & 6,867 &  8,584 & 25      \\
COVID-19 Tweets &  4,844 & 1,211 & 1,514  & 12   \\
\bottomrule
\end{tabular}
\label{tab:datasets}
\end{table}%

\subsection{Datasets}
We conducted a set of experiments on two public datasets provided by TREC~\cite{mccreadie2019trec}. 
Table~\ref{tab:datasets} gives an overview of each dataset: the number of tweets used in training (\textbf{\# Train}), validating (\textbf{\# Valid}), and testing (\textbf{\# Test}) our approach and baselines, in addition to the total number of classes (\textbf{\# Classes}).  
In particular, we split each dataset with $80\%-20\%$ ratio, where we use $80\%$ of tweets for training and $20\%$ for testing. During the training phase, we use $20\%$ from the training data to validate the model. We briefly summarize each dataset as follows:

\begin{itemize}
    \item \textbf{TREC-IS: }
    This dataset contains approximately $35$K tweets collected during $33$ different disasters between 2012 and 2019 (e.g., \texttt{wildfires, earthquakes, hurricanes, bombings,} and \texttt{floods}). 
	The tweets are labeled with $25$ information types by human experts and volunteers. 
	
	\item \textbf{COVID-19 Tweets: } This dataset contains a collection of tweets about the COVID-19 outbreak in different affected regions. In total, the data has $7,590$ tweets labeled with one or more of the full $12$ information type labels (the same as for the TREC-IS dataset).  
\end{itemize}

\begin{figure*}[t!]
\centering
    \subfigure[TREC-IS Dataset] 
    {
        \includegraphics[width=\columnwidth]{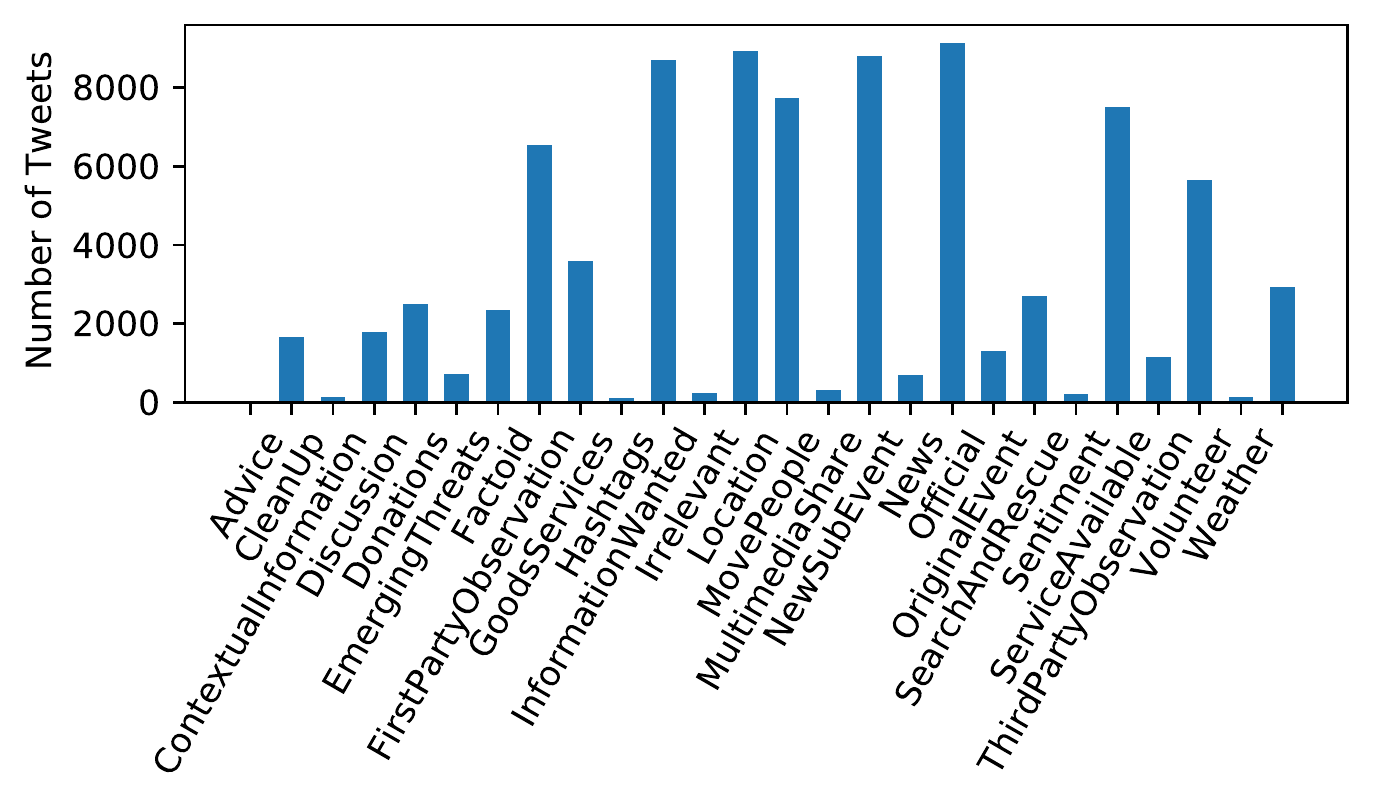}
    }
    \hfill
     \subfigure[COVID-19 Tweets Dataset]
    {
        \includegraphics[width=\columnwidth]{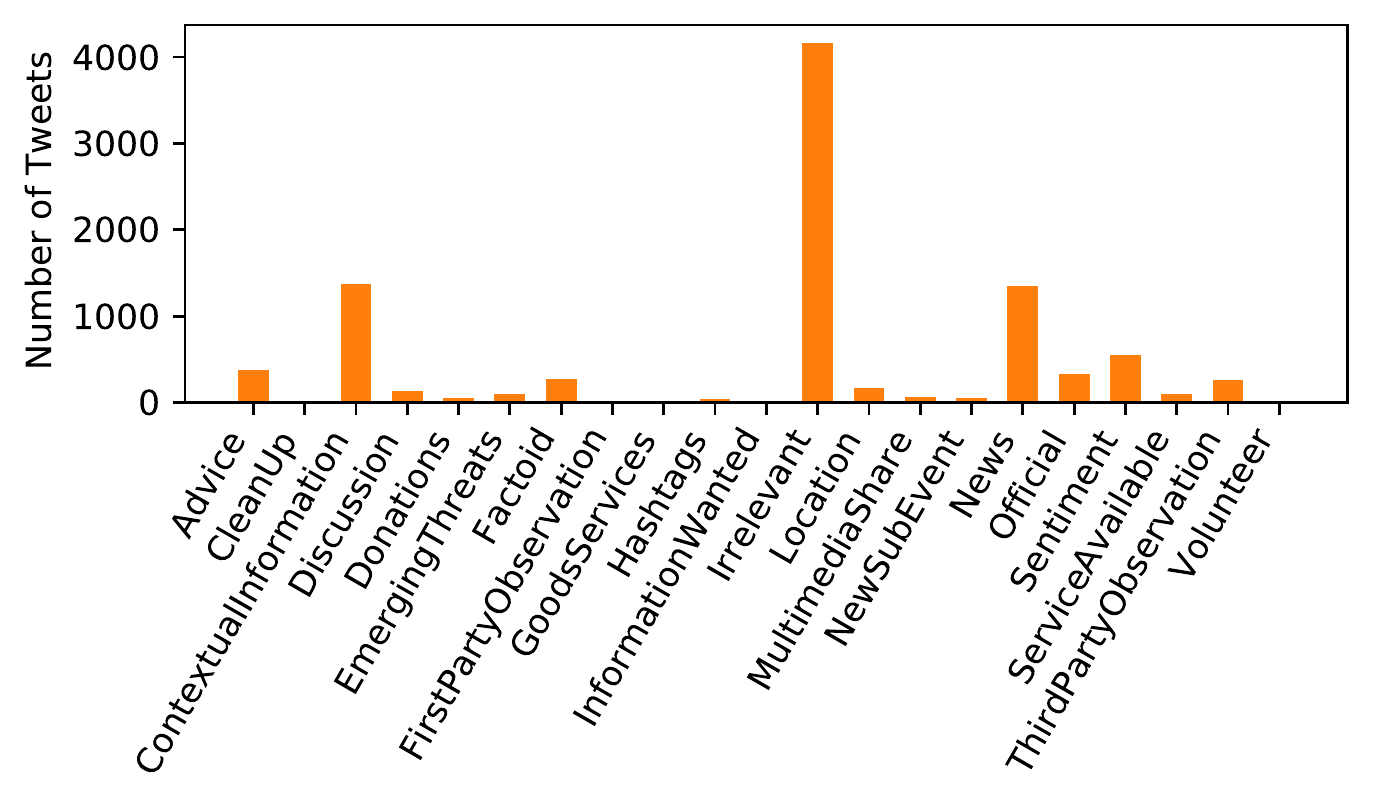}
    }   
\hfill    
\caption{Tweets' distribution across all information types in both datasets (TREC-IS and COVID-19 Tweets)}    
\label{fig:tweets-informationTypes}    
\end{figure*}

Figure~\ref{fig:tweets-informationTypes} shows the distribution of tweets per information type in both datasets. 
Apparently, the datasets are highly imbalanced \textit{w.r.t.} tweets' distribution across information types. 
For example, in the TREC-IS dataset, there are more than $6,000$ tweets that are categorized into the information types \texttt{Hashtags}, \texttt{News}, \texttt{MultimediaShare}, and \texttt{Location}. In contrast, the information types \texttt{CleanUP, InformationWanted}, and \texttt{MovePeople} have significantly fewer tweets. Similarly in COVID-19 Tweets, the tweets' distribution is extremely imbalanced: most tweets are categorized into \texttt{Irrelevant, ContextualInformation, Advice,} or \texttt{News}. This skewing distribution in tweets renders multi-label classification more challenging. 


	\subsection{Baselines} \label{sec:baselines}
	We consider a set of state-of-the-art approaches in multi-label classification\footnote{\scriptsize Note that we consider the relevant baselines that were implemented for multi-label classification. 
	The current version of VGCN-BERT cannot be directly used for multi-label classification. Therefore, we did not include it in the baselines.} as baselines in our evaluation. We briefly describe each baseline as follows:
	
\begin{itemize}
		\item \textbf{TextCNN}~\cite{kim2014convolutional} uses a convolutional neural network to construct text representation. 
		\item \textbf{HAN}~\cite{yang2016hierarchical} uses a hierarchical attention neural network to encode text with word-level attention on each sentence. 
		
		\item \textbf{BiLSTM}~\cite{zhou2016attention} is a bidirectional LSTM model that parses the text from left to right and right to left, then uses the final hidden state as a feature representation for the whole text.
		\item \textbf{MAGNET}~\cite{DBLP:conf/icaart/PalSS20} employs a bidirectional LSTM with BERT embeddings to represent tweets and GAT for labels classifiers. Then it uses a dot-product function to compute similarities between tweet vectors and labels' vectors. 
		
	\end{itemize}
	
\subsection{Implementation and Preprocessing}\label{sec:implementation}
We use the open-source implementations for TextCNN, HAN, and  BiLSTM models provided by the corresponding authors in their GitHub repositories. Furthermore, we implemented the code for the MAGNET model as it has not been open-sourced to date. In our approach, we use the implementation of BERT-Encoder from the \texttt{Huggingface}\footnote{\url{https://huggingface.co/}} library. 

Hyperparameters in the baselines are set with same values as mentioned in their original papers. In our model, we tune  hyperparameters via the grid search method to find optimal values for best performances. Specifically, our model achieves its best performance with the following values: \texttt{training-epochs} to $200$ with \texttt{batch-size} of $128$ and \texttt{Adam} optimizer~\cite{kingma2014adam} with a \texttt{learning-rate} of $2\mathrm{e}^{-5}$. To avoid overfitting, we add a \texttt{dropout} layer with a rate of $0.25$ and apply an \texttt{early-stopping} technique during model's training. The implementation of the I-AID model is open-sourced and available on the project website\footnote{\url{https://github.com/dice-group/I-AID}}.

\paragraph{Data Preprocessing} Given that the evaluation datasets are tweets, we perform ad-hoc preprocessing steps to capture the tweets' semantics. 
In particular, we perform the following preprocessing steps: 
(1) We use the \texttt{NLTK's TweetTokenize}\footnote{\url{https://www.nltk.org/api/nltk.tokenize.html}} API to tokenize tweets and retain the text content. 
(2) Stop-words, URLs, usernames, and Unicode-characters were removed. 
(3) Extra white spaces, repeated full stops, question marks, and exclamation marks are removed. 
(4) Emojis are converted to text using the \texttt{emoji}\footnote{\url{https://pypi.org/project/emoji/}} python library.
Finally, (5) \texttt{spaCy}\footnote{\url{https://github.com/explosion/spaCy}} library is used, to extract \texttt{named-entities} from tweets.

\subsection{Evaluation Metrics}\label{section:evaluation_metrics}
	We consider standard evaluation metrics for a multi-label classification task. In particular, we use a \textit{weighted average F1 score}, \textit{hamming loss} and \textit{Jaccard index} to evaluate the system's performance: 

	\begin{itemize}
	
	 \item{\textbf{Weighted average F1 score}}: 
	 F1 score is the harmonic mean of precision and recall scores. We use a \textit{weighted average} that calculates the F1 score for each label independently, then adds them together and uses a weight relative to the number of tweets in each label.
	 \begin{equation}
	    \text{F1}_{w.avg.}= 2 \sum_{i=1}^{k} \frac{|T_{\lambda_{i}}|}{|T|} \frac{precision_{\lambda_{i}}\times recall_{\lambda_{i}}}{precision_{\lambda_{i}} + recall_{\lambda_{i}}}
	 \end{equation}
	 
	 where $|T_{\lambda_{i}}|$ denotes the number of tweets with label $\lambda_{i}$ and $|T|$ is the total number of tweets. Precision$_{\lambda_{i}}$ and recall$_{\lambda_{i}}$ are the values of precision and recall for $\lambda_{i}$.
	 
	\item \textbf{Hamming Loss}: To estimate the error rate in classification, we use the \textit{hamming loss} function~\cite{schapire1999improved} 
	that computes the fraction of incorrectly predicted labels out of all predicted labels. Hence, the smaller the value, the better the performance.  
	
	\begin{equation}
	{hamming\:Loss\:}(y^{(i)}, \hat{y}^{(i)})=\frac{1}{S}\sum_{i=1}^{S}\frac{1}{k} |y^{(i)}\oplus\hat{y}^{(i)}|
	\end{equation}
 	where $S$ is the dataset size, $k$ is the total number of labels (i.e., $|\Lambda|$ ), $\oplus$ denotes the XOR operator, and $y^{(i)}$ and $\hat{y}^{(i)}$ are the groundtruth and predict labels, respectively, of tweet $i$ .
 	
	 \item \textbf{Jaccard Index:} To assess the system's accuracy, we use the Jaccard index to evaluate the similarity between predicted labels $\hat{y}^{(i)}$ and groundtruth labels $y^{(i)}$. 
	 Jaccard index computes the percentage of common labels in two sets of all labels as:
	\begin{equation}
	{jaccard\:}(y^{(i)}, \hat{y}^{(i)})=\frac{|y^{(i)} \cap \hat{y}^{(i)}|}{ |y^{(i)} \cup \hat{y}^{(i)}|}  
	\end{equation}
	where $y_i$ and $\hat{y}_i$ are the groundtruth and predicted labels for tweet $i$. $\cap$ and $\cup$ denote intersection and union set operations, respectively.
	\end{itemize}
	
	\subsubsection{Evaluating Actionable Information}
	We aim to evaluate the efficacy of our system in identifying tweets with actionable information,
	i.e., the system should trigger an alert if an input tweet includes actionable information (e.g., requests for search and rescue or reports of emerging threats). For this purpose, TREC-IS~\cite{mccreadie2020incident} introduces a new evaluation metric called \textit{Accumulated Alert Worth} (AAW) to evaluate systems in detecting actionable information during crisis. 
	The AAW score ranges from $-1$ to $+1$, where a positive value indicates highly critical information in a tweet while a negative score indicates it is less critical. More details about the AAW metric can be found in~\cite{TRECIS-AAW}. Here, we summarize the AAW metric as follows: 
	\begin{equation}
	\label{eq:AAW}
	AAW=
	\frac{1}{2}\sum_{t\in T}
	\begin{cases}
	\frac{1}{|T_h|} \cdot hPW(t) & \text{if } t \in T_h\\
	\frac{1}{|T_l|} \cdot lPW(t) &  \text{otherwise} 
	\end{cases}   
	\end{equation}
	
	where $T_h$ and $T_l$ denote the sets of tweets with high and low priorities, respectively.
	$hPW(t)$ is a scoring function for tweets that should generate alerts and $lPW(t)$ is a scoring function for tweets that should not generate alerts.
	Formally,
	
	\begin{equation*}
	    hPW(t)=
	    \begin{cases}
	\alpha+((1-\alpha)\cdot (\varphi(t)+\hat{\varphi}(t)) & \text{if } p_t^s>=0.7\\
	-1 &  \text{otherwise} 
	\end{cases}   
	\end{equation*}
	
	\begin{equation*}
	lPW(t)=    
	\begin{cases}
	argmax(-log(\frac{\delta}{2}+1), -1) & \text{if } p_t^s>=0.7\\
	\varphi(t)+\hat{\varphi}(t) &  \text{otherwise} 
	\end{cases}   
	\end{equation*}

	 where $p_t^s$ is the priority score of a tweet by the system, and $\varphi(t)$ and $\hat{\varphi}(t)$ are actionable and non-actionable scores, respectively, for tweet $t$,.

\subsection{Discussion}
\subsubsection{Performance Comparison ($Q_1$)}
		\begin{table*}[th!]
		\centering
		\caption{Evaluation results of our approach (I-AID) and baselines on two datasets: TREC-IS and COVID-19 Tweets using weighted average F1, Hamming Loss and Jaccard Index. Best results are in bold.}
		\begin{tabularx}{0.75\linewidth}{@{}p{2.2cm}p{2.1cm}p{1.5cm}p{1cm}p{1.5cm}p{1.1cm}XXXX@{}}
			\toprule
			\multirow{2}[2]{*}{\textbf{Datasets}} & \multirow{2}[2]{*}{\textbf{Metrics}}& \multicolumn{4}{c}{\textbf{Baselines}} & \multirow{2}[2]{*}{\textbf{I-AID}} \\ 
			\cmidrule(lr){3-6}
			{\textbf{}}&  & TextCNN & HAN & BiLSTM & MAGNET\\
			\cmidrule(r){1-1} \cmidrule(lr){2-2} \cmidrule(lr){3-6} \cmidrule(l){7-7}
			\multirow{3}{*}{TREC-IS} 
			& $\text{F1}_{w.avg.}$& 0.25  & 0.37 & 0.31 & 0.53 & \textbf{0.59}  \\
			& Jaccard Index & 0.18 & 0.28 & 0.19 & 0.38 & \textbf{0.43} \\
			& Hamming Loss & 0.24 & 0.15 & 0.26 & 0.09 & \textbf{0.07} \\
			
			\cmidrule(r){1-1} \cmidrule(lr){2-2} \cmidrule(lr){3-6} \cmidrule(l){7-7}
			\multirow{3}{*}{COVID-19 Tweets} 
			& $\text{F1}_{w.avg.}$ & 0.47  &  0.40 & 0.43  & 0.51 & \textbf{0.55}   \\
			& Jaccard Index &  0.33 & 0.28 & 0.21 & 0.40 & \textbf{0.43} \\
			& Hamming Loss &  0.11 & \textbf{0.04} & 0.07 &0.12 & 0.08 \\
			\bottomrule
		\end{tabularx}
		\label{tab:results}
	\end{table*}
	
	We use different metrics in multi-label classification to evaluate the performance of I-AID and baseline methods.  
	To ensure a fair evaluation, we use the same \textit{train} dataset for training all models and the \textit{test} dataset for evaluation. Table~\ref{tab:results} reports our evaluation results for each model on both datasets (TREC-IS and COVID-19 Tweets). 
	We consider the weighted average F1 score as the primary metric to compare and rank systems. Weighted average F1 takes into account the average performance of each system across all information types. 
	Overall, our approach (I-AID) achieves superior results to the other baselines under several metrics. In particular, our approach outperforms the weighted average F1 score of MAGNET---\textit{the state-of-the-art baseline in multi-label tweets classification}---by $+6\%$ on TREC-IS and $+4\%$ on COVID-19 Tweets. 
	
	We employ the Jaccard index and Hamming loss in further analysis to evaluate accuracy and error rate. 
	using Jaccard index, our approach outperforms all baseline methods in both datasets. In particular, I-AID achieves $43\%$ Jaccard index for both datasets compared with MAGNET's score of $38\%$ for the TREC-IS dataset and $40\%$ for COVID-19 Tweets. 
	On the other hand, our approach achieves suboptimal results using Hamming loss. For the TREC-IS dataset, I-AID achieves the best performance with rate $0.07\%$. While in COVID-19 Tweets, it achieves the second best score with $0.08\%$ compared with HAN model's score $0.04\%$. 
	
    Our experiments demonstrate that I-AID performs fairly well when categorizing disaster-related tweets into multiple information types. This is due to three facts:  
	i) we constructed a multimodel framework that leverages contextualized embeddings from the BERT model to capture contextual information in tweets. 
	ii) Our approach enriches the semantics of tweet representation by injecting label information and integrating additional structural information between tweets' tokens and labels using GAT.
	iii) Finally, we employ a Relation Network to \textit{learn} automatically similarities between tweets and labels. By using a learnable distance function, we learn an efficient metric in a supervised way to facilitate the mapping between a tweet and multi-label output.  
    
\subsubsection{Actionable Information In Tweets ($Q_2$)}
To answer $Q_2$, 
we use the AAW metric, proposed by TREC (Eq.~\ref{eq:AAW}), to evaluate the I-AID's ability to identify tweets with critical information. 
There are two ways to define an \textit{actionable} tweet~\cite{DBLP:journals/pacmhci/ZadeSRK0S18}: 
i) in terms of high priority information, commonly marked as critical by human assessors, and
ii) in terms of information type, for instance, a tweet with the labels \texttt{MovePeople} or \texttt{CleanUP} is considered more actionable than \texttt{News} or \texttt{Multimediashare}. In our evaluation, we consider the second definition of \textit{actionable} posts. 
The evaluation results of the AAW metric are presented in Table~\ref{tab:AAW}, where the top $6$ rows show the evaluation results for the baseline approaches in multi-label classification. 
The rest of Table~\ref{tab:AAW} shows the AAW results of the best approaches from the TREC-IS challenge (2019 edition~\cite{mccreadie2019trec} RUN B). The result of our approach (I-AID) is presented at the bottom of Table~\ref{tab:AAW}.
 
Our approach (I-AID) substantially outperforms all baseline approaches. 
In particular, in high priority AWW,  I-AID achieves an absolute improvement of $+26\%$ compared to the MAGNET model and $+32\%$ compared to \texttt{nyu-smap} (the best-achieved result in TREC-IS 2019). 
Furthermore, I-AID outperforms the Median score of TREC-IS participants by $+28\%$ in high priority and by $+30\%$ in overall AWW. 
Remarkably, our approach is the first to achieve a positive AAW score on high priority tweets. 
Although we outperform the state-of-the-art in both classification and AAW, the results of our evaluation suggest that a significant amount of research is still necessary to spot high priority tweets in a satisfactory manner. 
	
	\begin{table}[t!]
		\centering
		\small
		\caption{Performance evaluation using the AAW metric on the test dataset from TREC-IS (RUN B). A higher AAW value indicates better prediction.}
		\begin{tabularx}{0.95\linewidth}{@{}p{3cm}p{2.5cm}p{1.3cm}@{}}
			\toprule
			\multirow{2}[2]{*}{\textbf{Systems}} & \multicolumn{2}{l}{\textbf{Accumulated Alert Worth (AAW)}} \\ \cmidrule(r){2-3} 
			& High Priority             &       All      \\ \midrule
			
			TextCNN                                     &         -0.9764           &    -0.4884       \\ 
			HAN                                         &       -0.7816             &    -0.4600      \\
			Bi-LSTM                                      &      -0.8760              &    -0.4482     
			\\
			BERT (UPB\_BERT)  &         -0.9680         &       -0.4882     \\  
			TextGAT     &   -0.9794      &   -0.4897        \\
			MAGNET          &    -0.9436         &              -0.4726
			\\ 
			
			\midrule
			Median                                     &         -0.9197           &    -0.4609       \\
			BJUTDMS-run2                                &      -0.9942                  &   -0.4971 \\
			IRIT         &         -0.9942      &                   -0.4971  \\
			irlabISIBase &        -0.2337      &                -0.4935     \\
			UCDbaseline                                 &       -0.7856             &    -0.4131      \\
			nyu-smap &                  -0.1213 &               -0.1973 \\ 
			
			SC-KRun28482low &                   -0.9905 &                     -0.4955\\
			xgboost &                   -0.9942 &                                 -0.4972\\
			UCDrunEL2           &        -0.8556   &  -0.4382 \\
			cmu-rf-autothre &       -0.8481     &       -0.4456 \\
			\midrule
			I-AID                              &     \textbf{0.2044} &    \textbf{-0.1509}\\
			\bottomrule
		\end{tabularx}
		\label{tab:AAW}
	\end{table}
\subsubsection{Ablation Study ($Q_3$)}

\begin{table}[h!]
\centering
\caption{Ablation Study of I-AID Model}
		\begin{tabularx}{\columnwidth}{@{}p{2cm}p{1.8cm}p{1cm}p{1cm}llll@{}}
			\toprule
			{\textbf{Datasets}} & {\textbf{Metrics}} &  \textbf{I-AID-BERT} & \textbf{I-AID-TGAT} & \textbf{I-AID}\\
			\midrule
			\multirow{3}{*}{TREC-IS} 
			& $F1_{w.avg.}$ & 0.50  & 0.26 & 0.59\\
			& Jaccard Index & 0.34 & 0.18 & 0.43\\
			& Hamming Loss & 0.11 & 0.24 & 0.07\\
			\midrule
			\multirow{3}{*}{COVID-19 Tweets} 
			& $F1_{w.avg.}$ & 0.47  &  0.36 & 0.55\\
			& Jaccard Index &  0.37 & 0.15 & 0.43\\
			& Hamming Loss &  0.10 & 0.17 & 0.05\\
			\bottomrule
		\end{tabularx}
		\label{tab:ablationstudy}
	\end{table}

As discussed in Section~\ref{sec:approach}, our approach employs two main components (namely, BERT-Encoder and TextGAT) for representing input tweets. 
We perform an ablation study to evaluate the performance of each component individually. 
To do so, we implement two more versions of I-AID: the I-AID-BERT and the I-AID-TGAT. 
In the I-AID-BERT we deploy our system with the BERT-Encoder only to classify tweets into multiple information types. 
In the same manner, we implement I-AID-TGAT with only the TextGAT component. 
Table~\ref{tab:ablationstudy} shows the evaluation results for each component in our ablation study. 
Evidently, the BERT-Encoder-based implementation of I-AID achieves better performance than the TextGAT version. 
In particular, on the TREC-IS dataset, I-AID-BERT reach $50\%$ F1 score compared with $26\%$ by I-AID-TGAT. 
These results prove that I-AID-BERT can learn rich representation features from short text better than I-AID-TGAT. 
Moreover, in our approach we demonstrate that leverageing BERT and GAT together in a multimodel framework improves the overall performance. On the TREC-IS datastet, the I-AID model achieves superior results by $+9\%$ in F1 score compared with the BERT-based model and by $+33\%$ compared with the TGAT-based model. Our experiment on COVID-19 Tweets leads to a similar conclusion. Our approach outperforms BERT-based and GAT-based systems in F1 scores by $+8\%$, $+19\%$, respectively. On the other hand, we observe that the GAT-based model achieves better performance with fewer output labels. In COVID-19 Tweets with $12$ labels, I-AID-TGAT achieves an improved F1 score of $+10\%$ compared with its performance in TREC-IS with $25$ labels.

\section{Conclusion}\label{sec:conclusion}
In this paper, we propose I-AID, a multimodel approach for multi-label tweets classification. 
Our system combines three components: BERT-Encoder, TextGAT, and Relation Network. 
The BERT-Encoder is used to obtain locality information, while the TextGAT component aims to find correlations between tweets' tokens and their corresponding labels. 
Finally, we use a Relation Network as a last component output to learn the relevance of each label \textit{w.r.t.} the tweet content. 
Our main findings are as follows: 
i) Combining local information captured by BERT-Encoder and global information by TextGAT is beneficial for rich representation in short text and significantly advances multi-label classification. 
ii) Leveraging transfer learning from pretrained language models can efficiently handle sparsity and noise in social media data. 
iii) Benchmarking multi-label classification is a challenging task that requires proper evaluation metrics for fine-grained evaluation. 
I-AID achieves its best weighted average F-score of $0.59$ on the TREC-IS dataset. 
This result clearly indicates the sensitivity of our approach to the dataset's balancing. 
Dealing with unbalanced classes remains a future extension to our approach. We plan to use data augmentation and natural language generation to address this problem.

\bibliographystyle{IEEEtran}
\bibliography{references}	

\begin{IEEEbiography}[{\includegraphics[height=1.2in,clip]{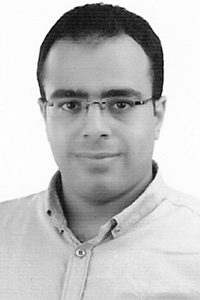}}]{Hamada M. Zahera} 
received the M.Sc. degree from the Computer Science Department, Faculty of Computers and Information, Menoufia
University, Egypt, in 2012. He is currently a PhD student at Data Science Group, University of Paderborn, Germany. His research interests include machine learning, Knowledge Graphs, and Semantic Computing. He act as a reviewer in ESWC, EACL conferences and PeerJ computer science journal.
\end{IEEEbiography}
\begin{IEEEbiography}[{\includegraphics[height=1.2in,clip]{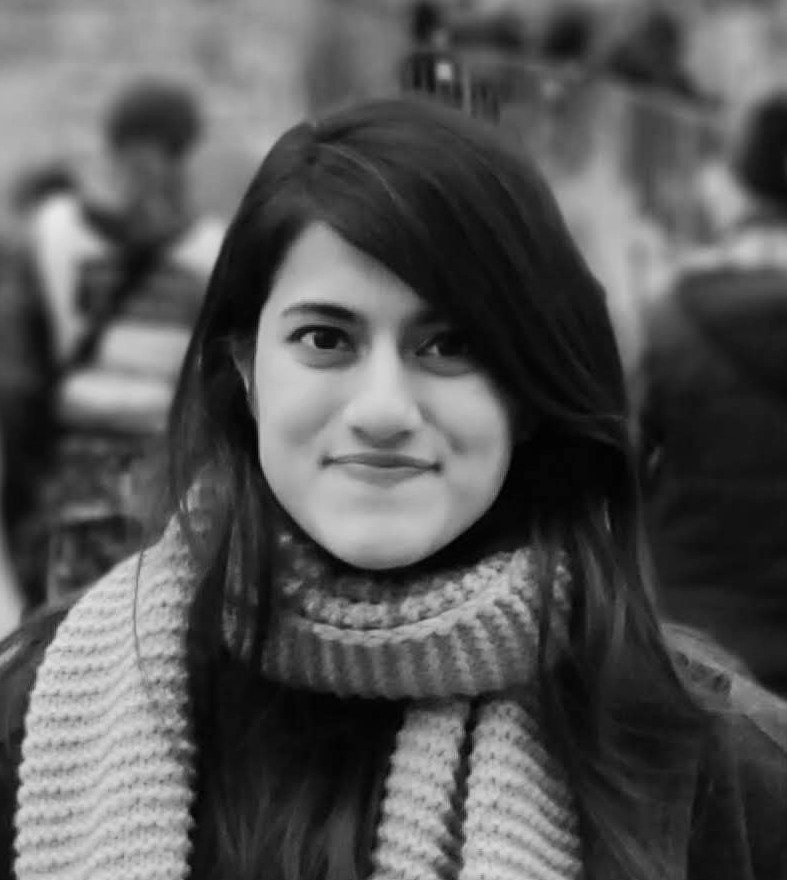}}]{Rricha Jalota} is a first-semester Master's in Computational Linguistics student at Saarland University and works as a student assistant in the Computer Science department of Paderborn University. Her interests lie in Language Representation, Reasoning, and Language Generation.

\end{IEEEbiography}
\begin{IEEEbiography}[{\includegraphics[height=1.2in,clip]{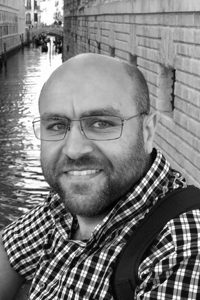}}]{Dr. Mohamed Ahmed Sherif}
is a postdoctoral researcher at the Data Science chair (DICE) at University of Paderborn. 
Mohamed's  research interests revolve around knowledge graphs and semantic web technologies, especially
(explainable) machine learning for data integration.
Mohamed developed number of algorithms for link specification learning, data repair, load balancing and relation discovery. 
Currently, he is leading the data integration tasks of many research projects.
\end{IEEEbiography}
\begin{IEEEbiography}[{\includegraphics[height=1.2in,clip]{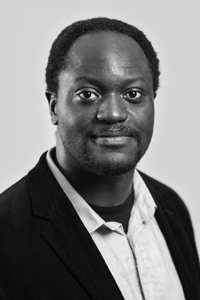}}]{Prof. Dr. Axel-Cyrille Ngonga Ngomo}
is the Data Science chair (DICE) at the Computer Science department at University of Paderborn. 
His research interests revolve around knowledge graphs and semantic web technologies, especially link discovery, federated queries, machine learning and natural-language processing. 
Axel has (co-)authored more than 200 reviewed publications and has developed several widely used frameworks.
\end{IEEEbiography}
\EOD
\end{document}